\documentclass[letterpaper, 10 pt, conference]{ieeeconf}  

\IEEEoverridecommandlockouts                              

\usepackage{graphics} 
\usepackage{epsfig} 
\usepackage{mathptmx} 
\usepackage{times} 
\usepackage{amsmath} 
\usepackage{amssymb}  
\usepackage{wrapfig}
\usepackage{comment}

\usepackage{hyperref}
\usepackage{url}            
\usepackage{booktabs}       
\usepackage{amsfonts}       
\usepackage{nicefrac}       
\usepackage{microtype}      
\usepackage{dsfont}
\usepackage[boxed]{algorithm2e}
\usepackage{graphicx} 
\usepackage{epstopdf}
\usepackage{xcolor}	
\usepackage{cite}

\usepackage{subcaption}
\usepackage{amsthm}
\theoremstyle{definition}

\makeatletter
\newcommand{\removelatexerror}{\let\@latex@error\@gobble}
\makeatother

\begin{document}

\title{Efficient Model Identification for Tensegrity Locomotion}

\author{Shaojun Zhu, David Surovik, Kostas Bekris and Abdeslam Boularias\\ Department of Computer Science, Rutgers University, New Jersey, USA\\ {\tt\small{ \{shaojun.zhu, david.surovik, kostas.bekris, abdeslam.boularias\}@cs.rutgers.edu}}}

\maketitle

\begin{abstract}
This paper aims to identify in a practical manner unknown physical
parameters, such as mechanical models of actuated robot links, which are critical in dynamical robotic tasks.  Key features include the use of an off-the-shelf physics engine
and the Bayesian optimization
framework. The task being considered is locomotion with a high-dimensional, compliant Tensegrity robot. A key insight, in this case, is the need to project the model identification challenge into an appropriate lower dimensional
space for efficiency. Comparisons with alternatives indicate that the proposed method can identify the parameters more accurately within the given time budget, which also results in more precise locomotion control. 

\end{abstract}

\section{INTRODUCTION}

This paper presents an approach for model identification by exploiting the availability of off-the-shelf physics engines used for simulating dynamics of robots and objects they interact with. There are many examples of popular physics engines that are becoming increasingly efficient \cite{ErezTT15, Bullet, MuJoCo, DART, PhysX, Havok}. These physics engines receive as input mechanical and mesh models of the robots in a particular scene, in addition to controls (force, torque, velocity, etc.) applied to them, and return a prediction of the robot's dynamical response. 

The accuracy of the prediction depends on several factors. The first one is the limitation of the mathematical model used by the engine (e.g., the Coulomb approximation). The second factor is the accuracy of the numerical algorithm used for solving the equations of motion. Finally, the prediction depends heavily on the accuracy of the physical parameters of the robots, such as mass, friction, and elasticity. In this work, we focus on the last factor and propose a method to improve the accuracy of the physical parameters used in the physics engine.

\begin{figure}
\centering
\includegraphics[width=0.4\textwidth]{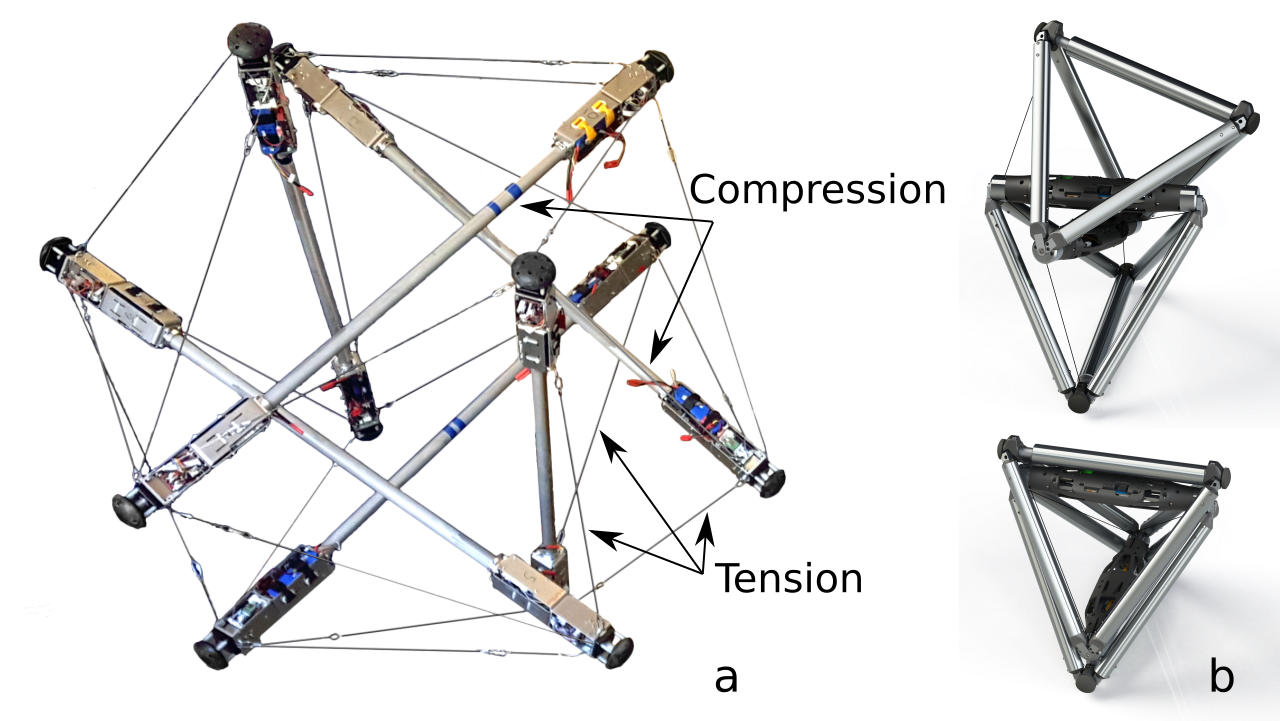}
\caption{ Tensegrity robots: a) NASA SUPERball:
a robotic icosahedron with 6 rods and 24 cables
\cite{sabelhaus2015system}. b) A duct climbing robot: 2 tetrahedral
frames with 8 actuated cables \cite{friesen2014ductt}. }
\label{fig:tensegrity}
\end{figure}

In the context of compliant locomotion systems, the Tensegrity robot of Figure \ref{fig:tensegrity} is a structurally compliant platform that can distribute forces into linear elements as pure compression or tension \cite{SunSpiral2014}. This robot's tensile elements can be actuated, enabling it to effectively adapt to complex contact dynamics in unstructured terrains. A policy for a rolling locomotive gait of the platform has been learned from simulated data \cite{DBLP:journals/corr/GengZBCVSAL16}. 

Tensegrity robots are inherently high-dimensional, highly-dynamic systems, and providing a predictive model requires a physics-based simulator \cite{NTRT}. The accuracy of such a solution critically depends upon physical parameters of the robot, such as the density of its rigid elements and the elasticity of the tensile elements. While a manual process can be followed to tune a simulation to match the behavior of a real prototype \cite{vytas_sim}, it is highly desirable to conduct this calibration using as few observed trajectories as possible. 

In this work, trajectories generated by a simulation manually tuned to a prototypical robotic platform are used to identify the parameters of a physics engine for tensegrity modeling. Given the high-dimensionality of the parameter space, this is a challenging problem. This work proposes the mapping of the model identification process to a lower dimensional space of parameters. Methods used for dimensionality reduction include Random Embedding (REMBO) \cite{wang2016bayesian} as well as Variational Auto Encoder (VAE) \cite{vae}. 

Furthermore, this work proposes to tie the dimensionality reduction process with the task performance by first learning a simplified dynamics model, then utilizing it to train an auto-encoder in the parameter space. Bayesian optimization is then conducted in the encoded space, avoiding much of the burden of high dimensionality. The proposed method is able to efficiently identify the parameters that produce a simulation that most closely matches the observed ground-truth trajectories of this exciting locomotive platform.

\section{Foundations and Contributions}
Two high-level approaches exist for learning robotic tasks with unknown dynamical models: model-free and model-based ones.  Model-free methods search for a policy that best solves the task without explicitly learning the system dynamics~\cite{Sutton:1998:IRL:551283, Bertsekas:1996, Bagnell_2013_7451, Levine_2014}. Model-free methods are accredited with the recent success stories of reinforcement learning in video games~\cite{mnih-dqn-2015}. For robot learning, a relative entropy policy search has been used~\cite{Peters+MA:2010} to successfully train a robot to play table tennis. The PoWER algorithm~\cite{kober2009policy} is another model-free policy search approach widely used in robotics. 

Model-free methods, however, do not easily generalize to unseen regions of the state-action space. To learn an effective policy, features of state-actions in learning and testing should be sampled from distributions that share the same support. This is rather dangerous in robotics, as poor performance in testing could lead to irreversible damage. 

Model-based approaches explicitly learn the dynamics of the system and search for an optimal policy using standard simulation, planning, and actuation control loops for the learned parameters. There are many examples of model-based approaches for robotic manipulation~\cite{Dogar_2012_7076,LunchMason1996,Mericli2014,isbell:physics:2014, ZhouPBM16}, some of which have used physics-based simulation to predict the effects of pushing flat objects on a smooth surface~\cite{Dogar_2012_7076}. A nonparametric approach was employed for learning the outcome of pushing large objects (furniture) ~\cite{Mericli2014}. A Markov Decision Process (MDP) has been applied to model interactions between objects; however, only simulation results on pushing were reported~\cite{isbell:physics:2014}. For general-purpose model-based reinforcement learning, the PILCO algorithm has been proven efficient in utilizing a small amount of data to learn dynamical models and optimal policies~\cite{Deisenroth:2011fu}.

Bayesian Optimization is a popular framework for data-efficient black-box optimization \cite{shahriari2016taking}. In robotics, some recent applications include learning controllers for bipedal locomotion \cite{antonova2016sample}, gait optimization~\cite{Calandra2016} and transfer policies from simulation to real world ~\cite{DBLP:conf/icra/MarcoBHS0ST17}.

Traditional system identification builds a dynamics model by
minimizing prediction error (e.g., using least
squares)~\cite{swevers1997optimal,Ljung:1999:SIT:293154}. There have
been attempts to combine parametric rigid body dynamics models with
nonparametric model learning for approximating the inverse
dynamics~\cite{nguyen2010using}. In contrast to such methods, this
work uses a physics engine, and concentrates on identifying mechanical
properties instead of learning the models from scratch. Recent work performed in simulation
only proposed model identification for predicting low dimensional physical parameters, such as either mass or friction~\cite{Yu-RSS-17}, before searching for an optimal policy.

This work is based on a model-based approach which utilizes a physics engine and concentrates on identifying only the mechanical properties of the objects instead of recreating the dynamics from scratch. Furthermore, it utilizes Bayesian optimization and identifies a dimensionality reduction process for dealing with high-dimensional model identification challenges efficiently.

\section{Model Identification}

This work proposes an online approach for robots to learn the physical parameters of their dynamics through minimal physical interaction. Because of the high dimensionality of the parameter space of the tensegrity robot, even very efficient methods such as Bayesian optimization (BO) struggle to identify all parameters with sufficient accuracy. 

This section introduces the overall framework of the model identification process. Dimensionality reduction methods, which decrease the search space of BO in order to achieve efficient optimization, are then covered in detail in the next section.

For the tensegrity robot, the physical properties of interest correspond to the density, length, radius, stiffness, damping factor, pre-tension, motor radius, motor friction, and motor inertia of the various rigid and tensile elements and actuators which are modeled in the NASA's Tensegrity Robotics Toolkit (NTRT) \cite{NTRT}. In total, 15 different parameters are considered.

These physical properties are represented as a $D$-dimensional vector $\theta \in \Theta$, where $\Theta$ is the space of all possible values of the physical properties. $\Theta$ is discretized with a regular grid resolution. The proposed approach returns a distribution $P$ on discretized $\Theta$ instead of a single point $\theta \in \Theta$.  This is appropriate due to the fact that model identification is generally an ill-posed problem, where multiple models can explain an observed trajectory with equal accuracy. The objective is to preserve all possible explanations for the purposes of robust planning.

The online model identification algorithm (given in Algorithm~\ref{greedyES_Alg}) takes as input a prior distribution $P_t$, for time-step $t\geq 0$, on the discretized space of physical properties $\Theta$. $P_t$ is calculated based on the initial distribution $P_0$ and a sequence of observations $(x_0,\mu_0, x_1,\mu_1, \dots, x_{t-1},\mu_{t-1}, x_{t})$. For the Tensegrity robot, $x_t$ is a state vector concatenating the 3D centers of all rigid elements, i.e., the rods in the corresponding Figure \ref{fig:tensegrity}, and $\mu_{t}$ is a vector of motor torques.

The process consists of simulating the effects of the controls $\mu_{i}$ on the robot in states $x_{i}$ under various values of parameters $\theta$ and observing the resulting states $\hat{x}_{i+1}$, for $i=0, \dots,t$. The goal is to identify the model parameters that make the outcomes $\hat{x}_{i+1}$ of the simulation as close as possible to the real observed outcome $x_{i+1}$.  In other terms, the following black-box optimization problem is solved: 
\begin{eqnarray}
\theta^* = \arg \min_{\theta \in \Theta} E(\theta)  \stackrel{def}{=}  \sum_{i=0}^{t}\|x_{i+1} - f(x_{i}, \mu_i, \theta) \|_2,
\label{gpError}
\end{eqnarray}
wherein $x_{i}$ and $x_{i+1}$ are the observed states of the robot at times $i$ and $i+1$,  $\mu_i$ is the control that applied at time $t$, and $f(x_{i}, \mu_i, \theta) =  \hat{x}_{i+1}$, the predicted state at time $t+1$ after simulating control $\mu_i$ at state $x_i$ using physical parameters $\theta$.

\begin{figure}
  \removelatexerror
\centering
\begin{minipage}[t]{0.5\textwidth}
  \begin{algorithm}[H]
\KwIn{State-action-state data $\{(x_{i},\mu_{i}, x_{i+1})\}$ for $i=0, \dots,t$\newline
 $\Theta$, a discretized space of possible values of physical properties\;}
\KwOut{Probability distribution $P$ over $\Theta$ according to the provided data\;}
Sample $\theta_0 \sim \textrm{Uniform}(\Theta)$; $L \leftarrow \emptyset$; $k\leftarrow 0$\;
\Repeat{ Timeout}{
$l_k \leftarrow 0$\;
\For{$i = 0$ \KwTo $t$}{
Simulate $\{(x_{i},\mu_{i})\}$ using a physics engine with physical parameters $\theta_k$ and get the predicted next state $\hat{x}_{i+1} = f(x_{i}, \mu_i, \theta_k)$ \;
$l_k \leftarrow l_k + \|\hat{x}_{i+1} - x_{i+1}\|_2$\;
}
$L \leftarrow L \cup \{(\theta_k , l_k)\}$\;
Calculate $GP(m,K)$ on error function $E$, where $E(\theta) = l$, using data $(\theta , l) \in L$\;

$\theta_{k+1} = \arg \max_{\theta\in \Theta} EI(\theta) $ \;
$k\leftarrow k+1$\;
}
\caption{\small {Model Identification with Bayesian Optimization}}
\label{greedyES_Alg}
\end{algorithm}
\end{minipage}
\end{figure}

The proposed approach consists of learning the error function $E$ from a sequence of simulations with different parameters $\theta_k\in \Theta$. To choose these parameters efficiently in a way that quickly leads to accurate parameter estimation, a belief about the actual error function is maintained. This belief is a probability measure over the space of all functions $ E: \mathbb{R}^D \rightarrow \mathbb{R}$, and is represented by a Gaussian Process (GP)~\cite{RasmussenGPM} with mean vector $m$ and covariance matrix $K$. The mean $m$ and covariance $K$ of the GP are learned from data points $\{\big(\theta_0, E(\theta_0)\big), \dots, \big(\theta_k, E(\theta_k)\big)\}$, where $\theta_k$ is a vector of physical properties of the object, and $E(\theta_k)$ is the accumulated distance between actual observed states and states that are obtained from simulation using $\theta_k$. High-fidelity simulations are computationally expensive. It is therefore important to minimize the number of simulations, i.e., evaluations of function $E$ while searching for the optimal parameters that solve Eq.~\ref{gpError}. BO decides the location for next sample by optimizing the acquisition
function. In our experiments, the expected improvement (EI) acquisition function
\cite{movckus1975bayesian} is used.

\section{Dimensionality reduction }

\subsection{Random Embedding for Model Identification}
\label{sec:random_embedding}

For problems where space $\Theta$ of physical properties has a high dimension $D$, the method presented in Algorithm~\ref{greedyES_Alg} is not practical because the number of elements in discretized $\Theta$ is exponential in dimension $D$.  This is a common problem in global search methods~\cite{wang2016bayesian}. In fact, it has been shown that Bayesian optimization techniques do not perform better than a random search when the dimension of the search space is too large (10 dimensions in the experiment in~\cite{AhmedNIPS2016}). Therefore, Algorithm~\ref{greedyES_Alg} cannot be directly used for robotic platforms with a large number of joints and parameters, such as the Tensegrity robot or compliant dexterous hands.


\begin{figure}
\centering
\hspace{-0.1in}
\includegraphics[width=0.5\textwidth]{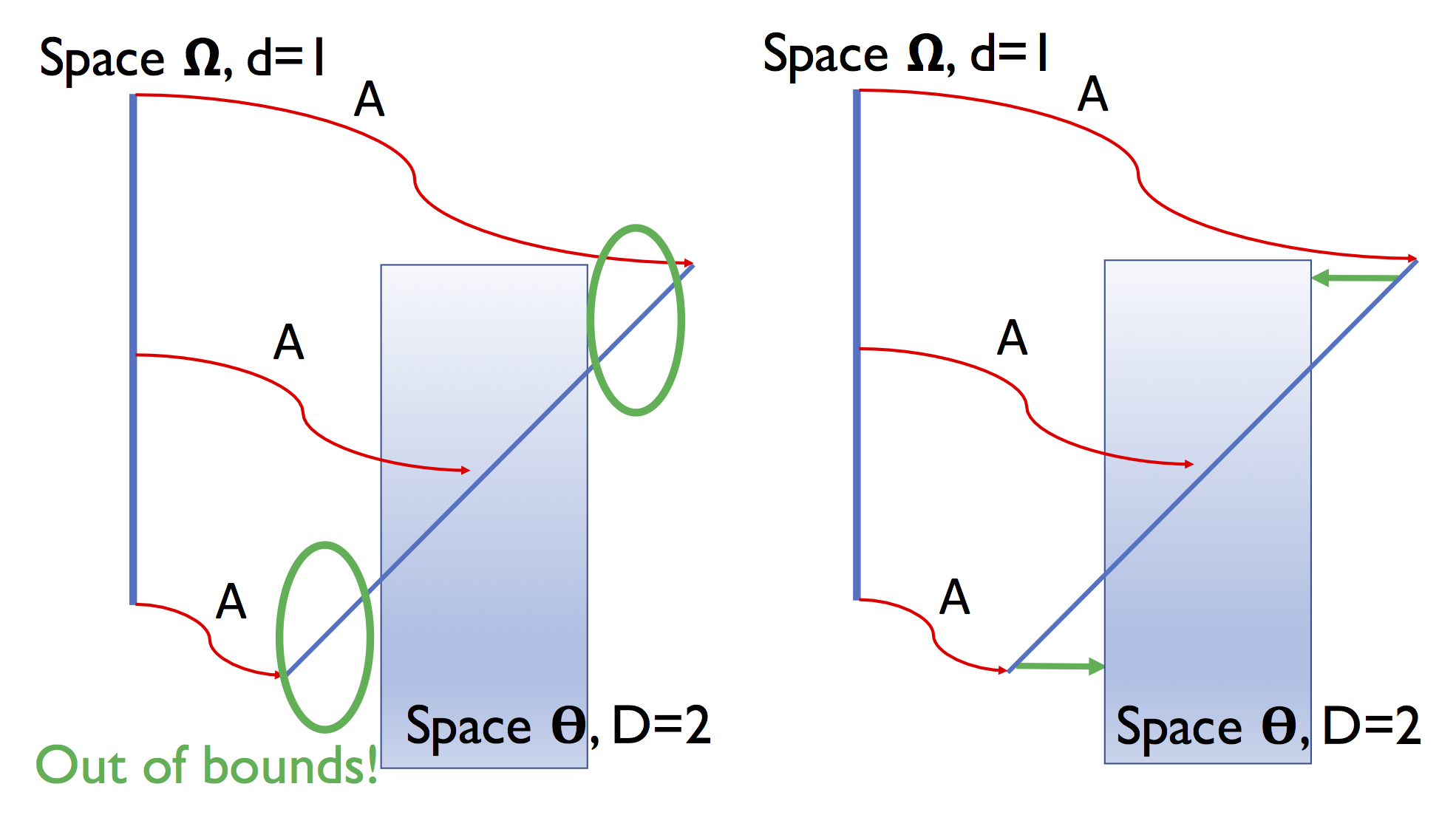}
\caption{LEFT: A example of 1D-to-2D projection resulting in points outside the original domain. RIGHT: REMBO resolves this issue by projecting the point outside $\Theta$ to the nearest boundary point of $\Theta$.}
\label{fig:tout_of_bound}
\end{figure}

Random embedding is an efficient and effective dimensionality reduction technique~\cite{wang2016bayesian}. Given a space of parameters $\Theta$ with dimension $D$, we generate a random matrix $A \in \mathcal{R}^{D \times d}$ that projects points from $\Theta\subset \mathbb{R}^D$ to a lower-dimensional space of parameters $\Omega\subset \mathbb{R}^d$ where $d<D$. Instead of discretizing $\Theta$, we discretize $\Omega$ into a regular grid and map each point $\omega \in \Omega$ to a point $\theta$ in the original high-dimensional space by using $A$, i.e. $\theta = A \omega$. One can show~\cite{wang2016bayesian} that with probability one, $\min_{\theta\in \Theta} E(\theta) = \min_{\omega\in \Omega} E(A\omega)$ where $E$ is the error function in Equation~\ref{gpError}. Consequently, we run Algorithm~\ref{greedyES_Alg} using discretized $\Omega$ as input instead of $\Theta$. We project back the low-dimensional vectors $\omega\in \Omega$ to original parameter space $\Theta$ using $\theta = A \omega$ when we need to run the physical simulation to get the trajectory under a sampled value of $\omega$. 

However, For a randomly generated matrix $A$ and point $\omega \in \Omega$, the corresponding high-dimensional vector $\theta = A\omega$ is not guaranteed to belong to $\Theta$, but could instead lie anywhere within $\mathbb{R}^D$. The simulator may consider $\theta$ as invalid if it is outside of $\Theta$ as shown in Fig.\ref{fig:tout_of_bound}. Moreover, simply using rejection sampling does not always work, since in some cases most of the sampled points will be invalid. {\it Random EMbedding Bayesian Optimization (REMBO) \cite{wang2016bayesian}} addresses this issue simply by projecting the point outside $\Theta$ to the nearest boundary point of $\Theta$.

\subsection{Variational Auto Encoder for Model Identification}
\label{sec:vae}

An autoencoder is a neural network that learns to reconstruct the input by going through a latent space, which is in a lower dimensional space than the original input space\cite{vincent2010stacked}. It has shown to be very useful in unsupervised learning of low dimensional representations. A variational autoencoder (VAE) adds an additional constraint that the latent space follows a prior distribution, usually assumed to be Gaussian \cite{vae}. This additional constraint makes the model more useful as a generative model, as it also learns to generate output from the prior distribution in addition to reconstruction.

\begin{figure}
\centering
\includegraphics[width=0.48\textwidth]{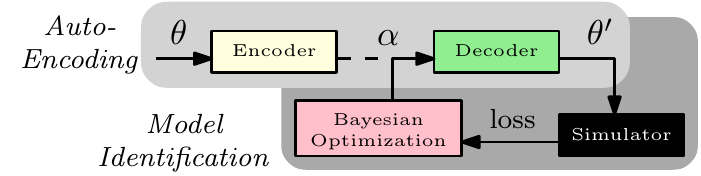}
\caption{The auto encoder is trained first to learn the latent low dimensional embedding. Then Bayesian optimization is performed in this low dimensional space to identify the optimal model parameter. The decoder is used to reconstruct the original 15 dimensional parameter in order to perform physical simulation.}
\label{fig:autoencodr_train}
\end{figure}

We adapt the VAE and combine it with the Bayesian optimization process, as shown in Fig. \ref{fig:autoencodr_train}. Firstly, the VAE is trained with randomly sampled physical parameter data $\theta$ to learn a low dimension embedding $\alpha$. Once the VAE is optimized, the decoder component is used to project the low dimensional $\alpha$ back to a value $\theta^\prime$ in the original physical parameter space. Thus, the Bayesian optimization process as detailed in Algorithm~\ref{greedyES_Alg} can be done efficiently in the low dimensional space. The decoder can be regarded as a learned non-linear version of the projection matrix $A$ in REMBO.

\subsection{Auto-Encoder with Learned Dynamics }
The use of VAE for reconstructing parameters from a low dimensional space has some limitations. Specifically, we are more interested in the accuracy of the predicted trajectory than in identifying the true underlying physical parameters. Mechanical models of motion can tie together several parameters of a model. Thus, connecting the dimensionality reduction process directly with the task performance may further improve the performance when using the identified model on the task. This idea is similar to learning a locally linear dynamics model while aiming to maximize the performance of the controller ~\cite{bansal2017goal}.


\begin{figure}[!ht]
\begin{center}
    \includegraphics[width=0.48\textwidth,trim={3mm 0mm 0mm 1.1mm},clip=true]{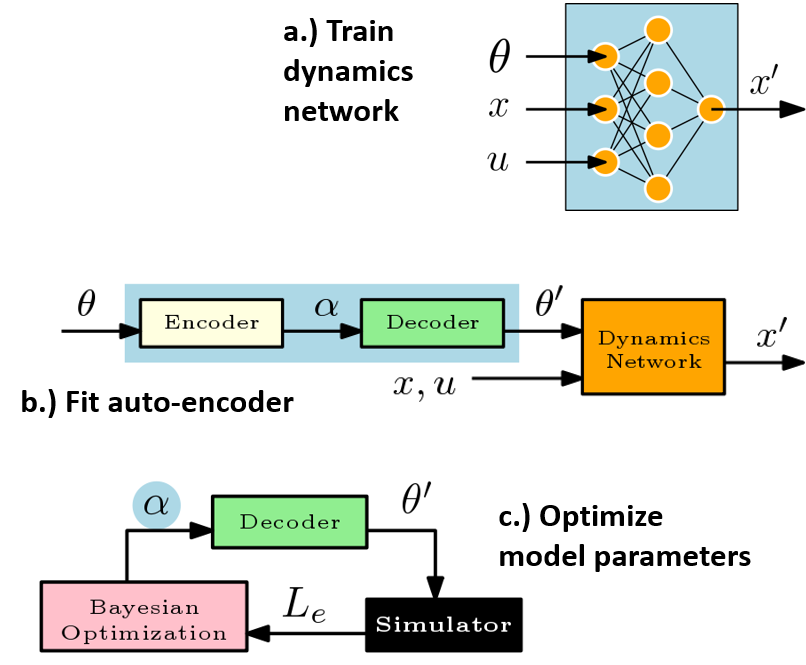}
\end{center}
\vspace{-.1in}
\caption{The three-step process for model identification under reduced parameter dimensionality. For each step, the element being updated is highlighted in blue.}
\vspace{-0.1in}
\label{fig:process}
\end{figure}

To provide intuition, we begin with an illustrative toy example of pushing a point-mass along a single dimension. We then show how the same approach can be applied to a much more complex system such as the Tensegrity robot.

\subsubsection{Toy example: Pushing}
\label{sec:push}
\begin{figure}[!ht]
\begin{center}
           \includegraphics[width=0.4\textwidth]{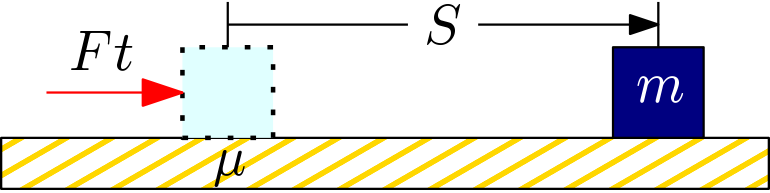}
\end{center}
\vspace{-.1in}
\caption{1-D point pushing example.}
\vspace{-0.1in}
\label{fig:push}
\end{figure}

Consider a cube of mass $m$ resting on a surface, as in Fig. \ref{fig:push}, which can be represented by a point that can only move along one axis. Assuming uniform and constant coefficient of kinetic friction $\mu$ between the cube and its resting surface, an impulse $Ft$ is applied to the cube to cause displacement across some distance $S$. Applying Newton's Laws and Kinematic Equations, we have the following equations:
\begin{align}
    Ft &= \mu m t_1 = m v_0 \\
    S &= v_0 t_1 /2 
\end{align}
where $v_0$ is the initial velocity of the cube after the impulse and $t_1$ is the duration of the cube's movement. Solving the above equations, we have:

\begin{equation}
S(m, \mu) = (Ft)^2/(2m^2\mu)
\label{push_eq}
\end{equation}

We use this equation solely to generate training and testing data, in the same manner as a black-box simulator. This parallels later use of a physics simulator for the Tensegrity robot without direct exposure of the differential equations of motion. Our goal is to identify $m$ and $\mu$, given only the initial impulse $Ft$ and the displacement $S$, without mathematical analysis of the system dynamics equations. 

This problem, however, is ill-posed due to the fact that different values of $m$ and $\mu$ can result in the same $S$ for a given value of $Ft$. For example, assuming $Ft=1$, both $m=1, \mu=0.32$ and $m=0.8, \mu = 0.5$ will result in $S = 1.5625$. In other words, as long as the value of $m^2\mu$ is uniquely identified, so is the displacement $S$. Thus, if the task is to predict $S$, it is not necessary to individually identify $m$ and $\mu$; the scalar value $m^2\mu$ can still uniquely determine the system. The goal then is to identify this one-dimensional representation automatically.


 Firstly, we train a dynamics network to predict the displacement $S$ given inputs $m$, $\mu$, and $Ft$, as shown in Fig. \ref{fig:process}a. In this case $\theta = {[m, \mu]}$ and $u=Ft$. The input state $x$ is omitted as we assume the point is always at the origin before being pushed. Secondly, we use the resulting dynamics network to train the auto-encoder to reconstruct $m$ and $\mu$, as shown in Fig. \ref{fig:process}b. During this step, the weights in the dynamics network are fixed. The encoder module is designed to receive input $m$, $\mu$ and output one-dimensional $\alpha$, which is influenced by the previous observation. A unique aspect of the auto-encoder is that, instead of using reconstruction error of $m$ and $\mu$ as the loss function, it passes the reconstructed parameters to the dynamics network and uses the displacement error. The goal of the auto-encoder is to reconstruct parameters resulting in similar dynamics, as predicting dynamics is the primary concern. 

%
\subsubsection{Tensegrity Robot}


This procedure is next applied to the much more complex Tensegrity robot system. Here, we assume the existence of a low dimensional representation of both the physical parameter space and the state-action space that, once identified, can determine the system dynamics similar to $m^2\mu$ in Sec. \ref{sec:push}. 

One challenge in adapting the procedure is that the Tensegrity robot is inherently a high-dimensional, highly-dynamic system which makes learning a dynamics model extremely difficult. Instead, a simplified dynamics model can be learned where instead of using the full state which is 126 dimensional, only the height of the center of the mass of the robot is used as the state.\footnote{The selection of the state representation is not a focus of the paper. The search for optimal state representation is left for future work.} In experiments, using the full state as input state and the height of the center of the mass as output results in better accuracy than using only the height of the center of the mass as both input and output. 

Thus, the simplified dynamics model takes parameters, full state, and action as input and predict the height of the center of the mass of the next time step, as shown in Fig. \ref{fig:process}a. In this case, $\theta$ is the 15-dimensional parameter, $x$ is the 126-dimensional full state, $u$ is the 24-dimensional action and $x'$ is the 1-dimensional height of the center of the mass of the robot. Using the simplified dynamics model, we train an auto-encoder on top of it to learn the low dimensional presentation of the parameters.

Similar to the 1D pushing example, the auto-encoder in Fig. \ref{fig:process}b is trained using loss function of the error between the new state simulated using original parameter $\theta$ and the new state predicted by the dynamics model using the reconstructed parameter $\theta'$. Afterward, Bayesian Optimization is performed using the decoder, as shown in Fig. \ref{fig:process}c.

\section{Experimental Results}
\label{sec:result}
\subsection{Toy example: 1-D Point Pushing}
\vspace{-0.2in}
\begin{figure}[!ht]
\begin{center}
         \includegraphics[width=0.4\textwidth]{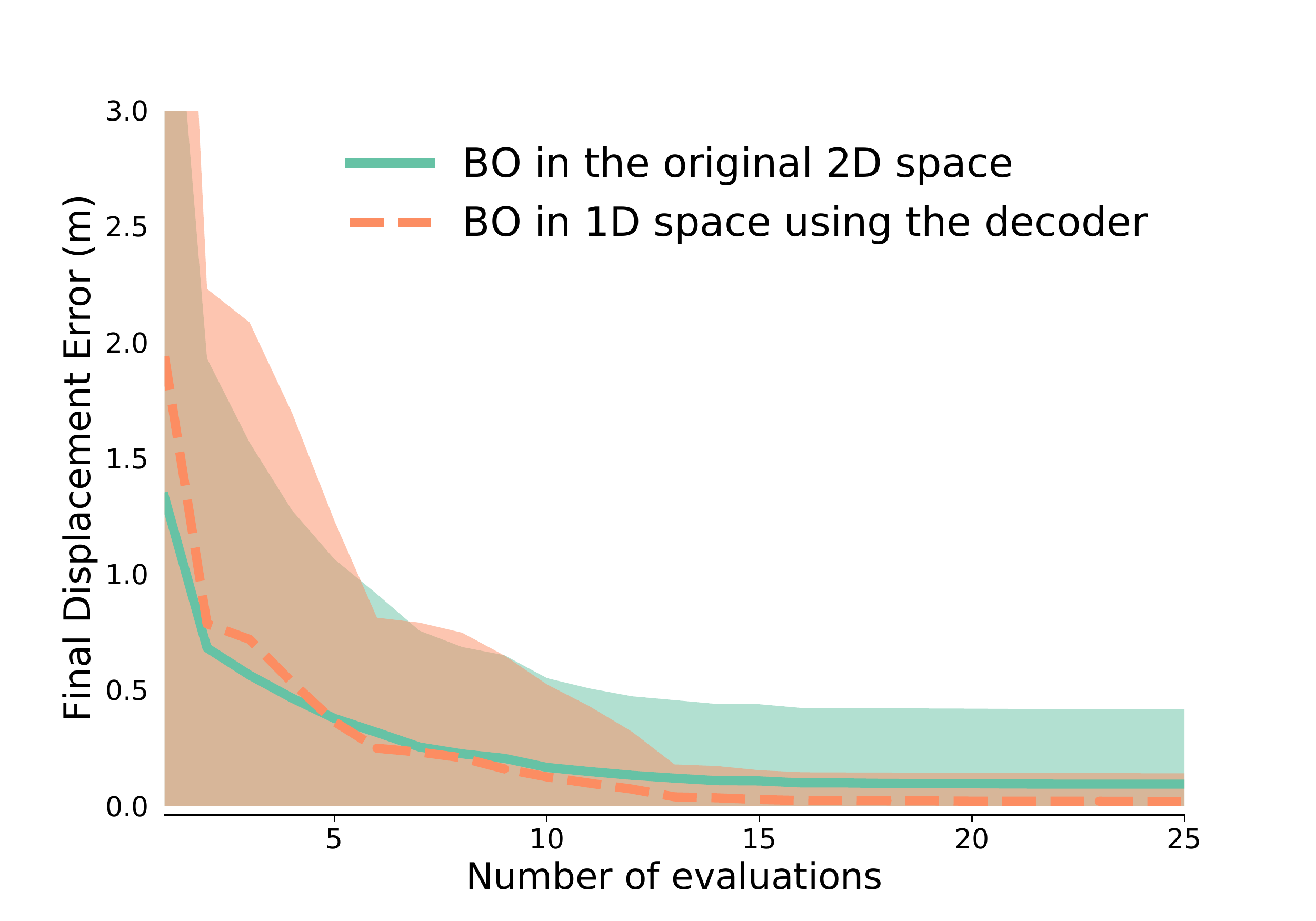}
           \caption{1-D point pushing result: Identifying model in the 1D space is more efficient than in the original 2D parameter space.} 
\label{fig:push_curve}
\end{center}
\end{figure}

In this toy example, we use equation \ref{push_eq} as the "simulator" to generate data. 20,000 training data points are generated by randomly sampling $m$, $\mu$, $F$ and $t$. 2,000 additional data points are generated for testing. As a proof of concept, we only compare model identification using Bayesian optimization in the original 2D parameter space and in the 1D latent space in the auto-encoder.

To train the dynamics network and the auto-encoder, The dynamics network has three hidden layers with 64, 128, and 64 hidden units and ReLU activation functions. The encoder network has two layers with 32, 1 units, and decoder has two layers with 1, 32  units. Both the encoder and the decoder only have ReLU on the first hidden layer.

The result is shown in Fig. \ref{fig:push_curve}. After training, the decoder is able to reconstruct $m$ and $\mu$ which results in close to zero error in the final displacement. Comparing to optimizing in the original 2D parameter space, using the decoder to optimize in 1D space is both more efficient and achieves lower error.
\subsection{Tensegrity Robot}

\begin{figure}
\centering
\includegraphics[width=0.4\textwidth]{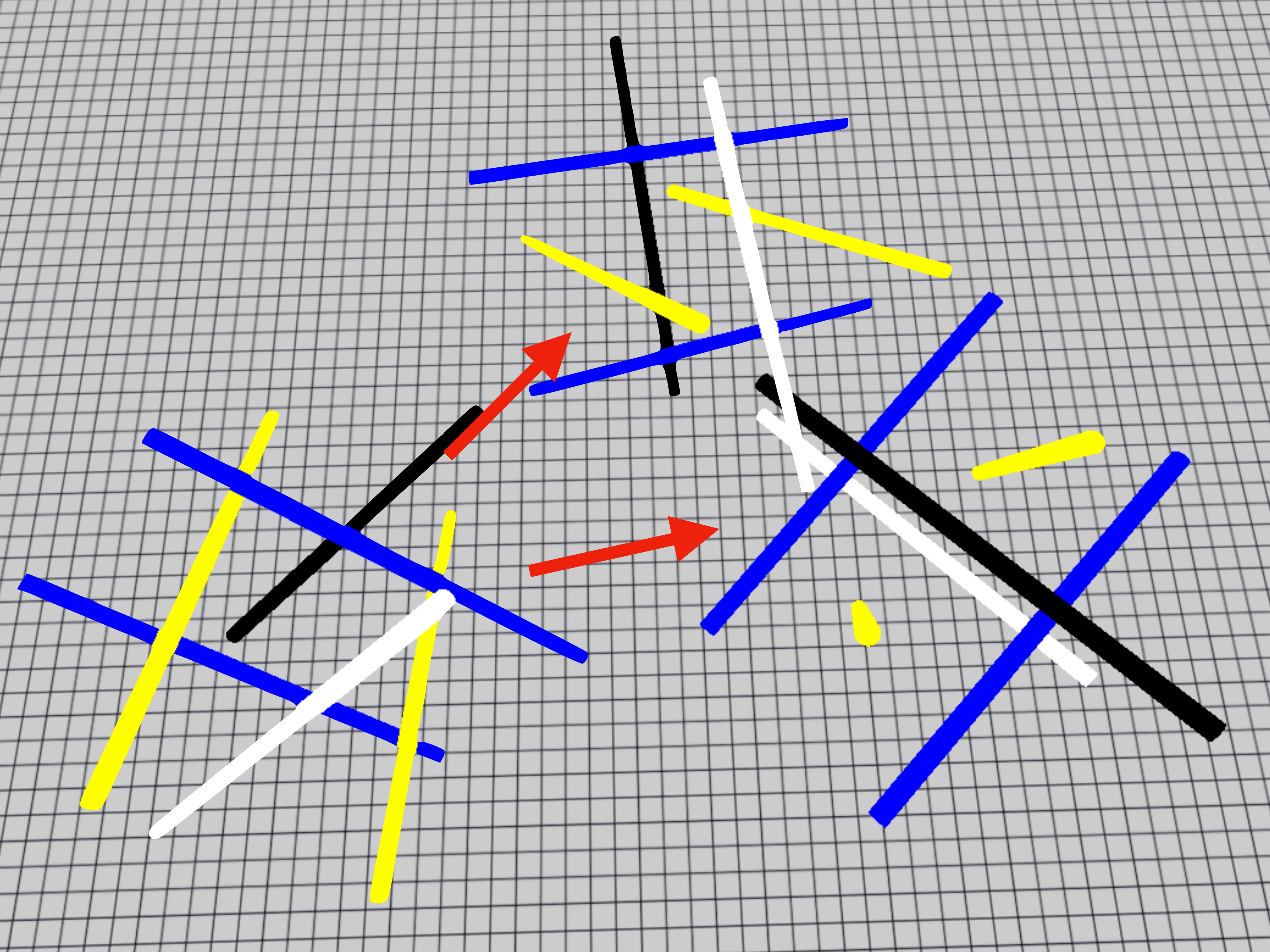}
\caption{Simulation of the Tensegrity robot resulting in different states when executing the same control for different robot model parameters.}
\label{fig:tensegrity_sim}
\end{figure}

\noindent {\bf Setup:} This experiment aims to identify the 15 parameters of the model of the Tensegrity SuperBall robot in NASA's Tensegrity Robotics Toolkit \cite{NTRT}. The complex dynamics and high dimensionality of the robot make this problem very hard. Fig. \ref{fig:tensegrity_sim} shows an example of the different results of applying the same control to the robot with 1\% difference in the rod length (one of the 15 parameters). In absence of access to the real robot, the default values of the T6 model in NTRT are used as ground-truth. 

The applied control law conducts an optimization procedure on the system's geometric configuration alone, without accounting for dynamics~\cite{isrr_control}.
Under the assumption that the base triangle remains in full contact with the ground, this law commands a change in cable lengths that correspond to a desired shift in the system's center-of-mass. By displacing this value relative to the supporting base triangle, the system can be made dynamically unstable, causing a forward flop.

By using the controller above, 1200 trajectories are generated as training data by sampling the 15 parameters within the $\pm 10\%$ range of the ground truth, using the NTRT simulator. The assumption is that such error can appear during the robot modeling process and the proposed approach should be able to minimize such error. Examples of the trajectories can be found on \url{https://sites.google.com/view/tensegrity/}. 

Random search is compared against as a baseline, where random values of the parameters are selected within the $\pm 10\%$ range. Nevertheless, it is well-known that Bayesian optimization in high dimensions is difficult due to the exponential growth of the search space. To deal with this issue, the two dimensionality reduction methods, REMBO and VAE are used to reduce the dimensionality of the parameter space from 15 to 5\footnote{The selection of the optimal low dimension is left for future work.}. 

The encoder and decoder of the VAE used in the experiment are both two-layer neural networks. The input dimension of the encoder and the output dimension of the decoder is 15, which is the dimension of the parameter space. The latent space is 5 dimensional. Between them is one layer of 400 dimensions. This dimension is chosen through cross-validation by balancing accuracy and network complexity. The prior distribution of the latent space in the VAE is assumed to be $N(0, 1)$. Based on the three-sigma rule, when sampling between $[-3, 3]$, this interval should cover $99.7\%$ of the latent space when the VAE is optimized. For REMBO, each time a random projection matrix is generated to project the parameters into $[0, 1]$.

The dynamics network is a four-layer neural network with dimensions 128, 64, 32, and 1. The encoder and decoder with the dynamics network are also two-layer neural networks but much narrower than the VAE with only 10 and 5 dimensions for each layer.

\begin{figure}
\centering
\includegraphics[width=0.49\textwidth]{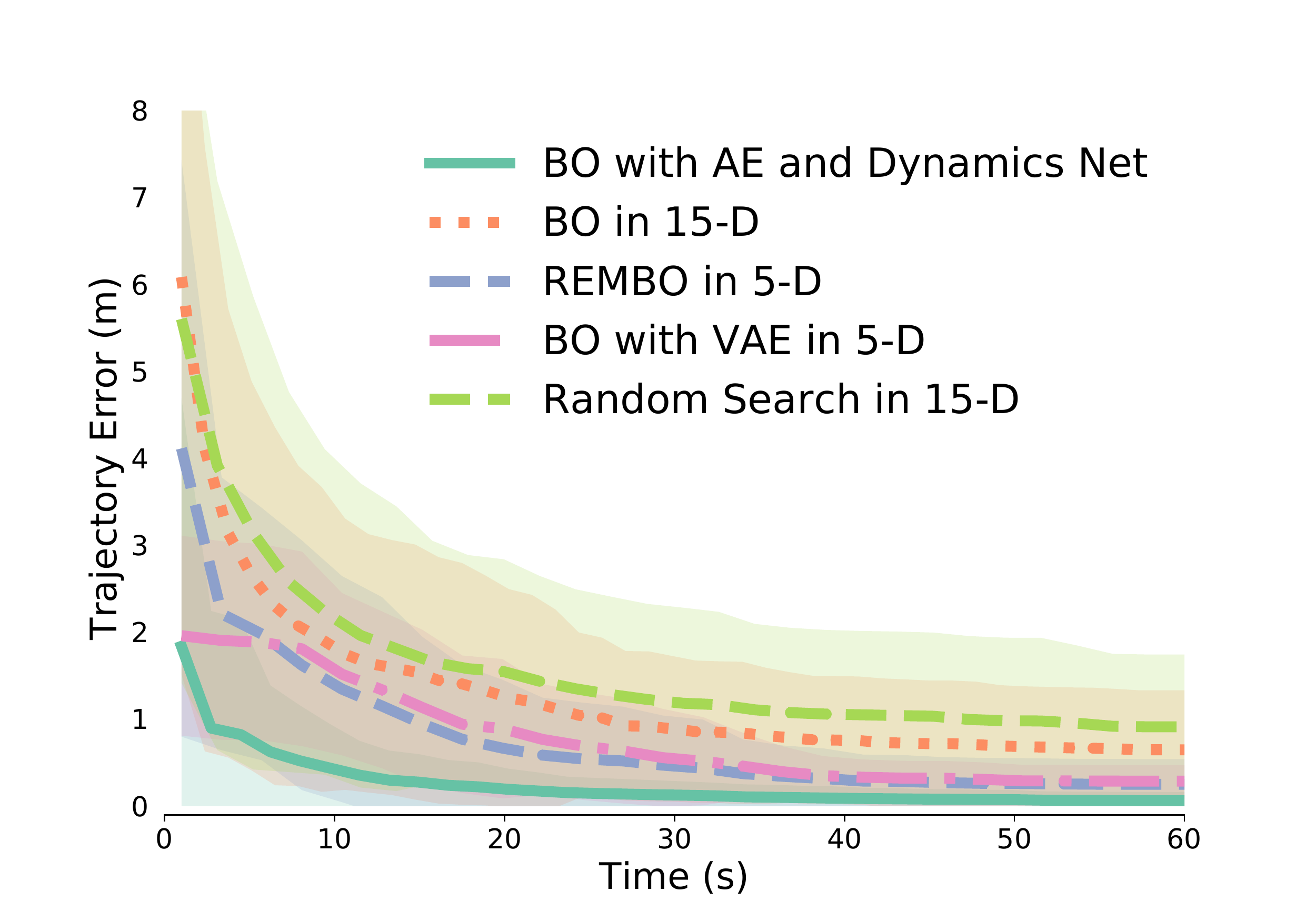}
\caption{Test trajectory errors of different methods for the Tensegrity robot as a function of time budget for the parameter optimization process. Bayesian optimization with the autoencoder with the learned dynamics network in the 5-dimensional space achieves the lowest trajectory error, outperforming random search and Bayesian optimization in the original 15 dimensional space, as well as REMBO or VAE in the 5-dimensional space. }
\label{fig:tensegrity_traj_error}
\end{figure}

\noindent {\bf Results:} Fig. \ref{fig:tensegrity_traj_error} shows the average error between the trajectories using the model parameters identified by different methods and the trajectories generated from the ground-truth simulator. When optimizing in the original 15-dim. space, as a data-efficient global optimization method, Bayesian optimization outperformed random search. Further improvements are achieved by dimensionality reduction, making the search more efficient. Bayesian optimization with the autoencoder with the learned dynamics network (BO with AE and Dynamics Net) in the 5-dimensional space achieves the lowest trajectory error, outperforming the method using REMBO or VAE. This shows that a learned better latent embedding enables more efficient parameter search in the Bayesian optimization process. 

Table \ref{tbl:tensegrity_prop_error} provides the errors for each of the final identified parameter. Interestingly, although achieved lowest trajectory error, BO with AE and Dynamics Net did not identify all parameters with lowest error. Specifically, it turns out that parameters like rod\_length, rod\_space, rod\_length\_mp, motor\_radius, motor\_friction, and motor\_inertia are not actually used in the current model of the SuperBall simulation. Thus even methods like VAE may be able to get lower reconstructing error on the parameters themselves, BO with AE and Dynamics Net is able to achieve lower trajectory error as it ties the model identification process with the dynamics. Additionally, some parameters may have a stronger influence on the robot dynamics. An intelligent way to identify these parameters would be helpful to reduce the dimensionality of the parameter space and could be more informative than random embeddings. This will be a direction for future work.

\begin{table*}[]
\centering
\caption{Identified Prameter Error (\%)}
\label{tbl:tensegrity_prop_error}
\begin{tabular}{l|l|l|l|l|l}
                & Random Search in 15-D & BO in 15-D & REMBO in 5-D & BO with VAE in 5-D & BO with AE and Dynamics Net \\\hline
density         & 5.01$\pm$2.86 & $4.10\pm2.66$ &     1.85$\pm$1.88 & $1.96\pm0.46$ &  \textbf{1.30$\pm0.05$}  \\\hline
radius          & 2.49$\pm$1.94  & $2.08\pm1.73$ &     1.86$\pm$ 1.84 & 1.43$\pm0.28$  &\textbf{0.30$\pm0.03$} \\\hline
density\_mp     & 5.40$\pm$2.96  & $5.19\pm2.66$ &      1.89$\pm$   1.86     & 2.38$\pm0.39$  &\textbf{1.00}$\pm0.18$   \\\hline
radius\_mp      & 4.78$\pm$2.78  & $5.36\pm2.97$ &        1.94$\pm$  1.94     &       2.00$\pm0.55$   &\textbf{0.69}$\pm0.43$    \\\hline
stiffnessActive & 4.49$\pm$2.68  & $4.44\pm2.79$  &   1.84$\pm$       1.90     & \textbf{1.68$\pm$0.46}     &1.71$\pm0.03$     \\\hline
damping         & 4.62$\pm$2.75  & $4.33\pm2.78$ &   1.81$\pm$    1.89        &    \textbf{2.02$\pm$0.44}        &2.26$\pm0.15$ \\\hline
rod\_length     & 5.05$\pm$2.75  & $4.72\pm2.69$ &  \textbf{1.90}$\pm$ 1.88          &  2.04$\pm$0.31 &6.25$\pm$0.59          \\\hline
rod\_space      & 4.96 $\pm$2.81  & $4.87\pm2.74$ &  1.88$\pm$   1.84          &      \textbf{1.68}$\pm$0.36               &5.20$\pm0.38$ \\\hline
rod\_length\_mp & 4.89 $\pm$ 2.81 & $5.22\pm2.87$ & 1.88$\pm$      1.96        &     \textbf{1.70}$\pm$0.58                &4.06$\pm0.23$ \\\hline
pretension      & 5.10 $\pm$ 2.83  & $5.10\pm3.01$ & 1.93$\pm$   1.89           &    1.58$\pm$0.50       &\textbf{1.38$\pm0.34$} \\\hline
maxTens         & 4.99 $\pm$ 2.87  & $4.66\pm2.80$ &  1.86$\pm$   1.83          &      \textbf{1.85}$\pm$0.42               &5.48$\pm0.12$\\\hline
targetVelocity  & 4.85 $\pm$ 2.62  & $5.31\pm3.01$ &  1.84$\pm$  1.90           &     2.06$\pm$0.62                &\textbf{0.49$\pm0.31$}                              \\\hline
motor\_radius   &   5.11      $\pm$ 2.90     & $4.38\pm2.81$ &  1.90$\pm$ 1.91            &    \textbf{1.79}$\pm$0.66                 &4.9$\pm0.23$                              \\\hline
motor\_friction & 5.10       $\pm$ 2.71   & $5.32\pm3.17$ & 1.89$\pm$ 1.82             &    2.19$\pm$0.27                 &\textbf{0.65$\pm0.03$}                              \\\hline
motor\_inertia  & 4.78    $\pm$ 2.80  & $4.87\pm3.01$ &  1.83$\pm$ 1.88            &     2.00$\pm$0.45                &\textbf{1.86$\pm0.06$}                             \\
\hline
\end{tabular}
\end{table*}

\section{Conclusion}
\label{sec:conclusion}

This work proposes an information and data efficient framework for identifying physical parameters critical for robotic tasks, such as compliant robot locomotion. The framework aims to minimize the error between trajectories observed in experiments and those generated by a physics engine. To solve high-dimensional challenges, this work integrates Bayesian optimization with a projection to a lower-dimensional space through random embedding or learning a latent embedding utilizing auto encoder. The evaluation of the proposed method against alternatives is favorable both in terms of identifying parameters more efficiently, as well as resulting in more accurate locomotion trajectories. 

An interesting extension of this work would involve the identification of controls during the learning process that help in quickly minimizing the error. This can be a robust control process, which takes advantage of Bayesian Optimization's output in terms of a belief distribution for the identified parameters, so as to minimize entropy and maximize the safety of the experimentation process. Furthermore, it is interesting to compare the generality of the learned models and resulting control schemes that utilize them against completely model-free and end-to-end approaches for reinforcement learning and control.

\bibliographystyle{IEEEtran}
\bibliography{../bib/model_based_rl}

\end{document}